# The Paradigm Shifts in Artificial Intelligence

Vasant Dhar[1]

July 2023

**Abstract.** Kuhn's framework of scientific progress (Kuhn, 1962) provides a useful framing of the paradigm shifts that have occurred in Artificial Intelligence over the last 60 years. The framework is also useful in understanding what is arguably a new paradigm shift in AI, signaled by the emergence of large pre-trained systems such as GPT-3, on which conversational agents such as ChatGPT are based. Such systems make intelligence a commoditized general purpose technology that is configurable to applications. In this paper, I summarize the forces that led to the rise and fall of each paradigm, and discuss the pressing issues and risks associated with the current paradigm shift in AI.

## 1. Introduction

Artificial Intelligence (AI) captured the world's attention in 2023 with the emergence of pre-trained models such as GPT-3, on which the conversational AI system ChatGPT is based. For the first time, we can converse with an entity, however imperfectly, about anything, as we do with humans. This new capability provided by pre-trained models has created a paradigm shift in AI, transforming it from an application to a general purpose technology that is configurable to specific uses. Whereas historically an AI model was trained to do one thing well, it is now usable for a variety of tasks such as general conversations, assistance, decision-making, or code generation – for which it wasn't explicitly trained. The scientific history of AI provides a backdrop for evaluating and discussing the capabilities and limitations of this new technology, and the challenging that lie ahead.

Economics Nobel Laureate Herbert Simon, one of the fathers of Artificial Intelligence, described Artificial Intelligence as a "science of the artificial." (Simon, 1970). In contrast to the natural sciences, which describe the world as it exists, a science of the artificial is driven by a goal, of creating machine intelligence. According to Simon, this made AI a science of design and engineering. Pre-trained models have greatly expanded the design aspirations of AI, from crafting high performing systems in narrowly-specified applications, to becoming general-purpose and without boundaries, applicable to anything involving intelligence.

The evolution of AI can be understood through Kuhn's (1962) theory of scientific progress in terms of "paradigm shifts." A paradigm is essentially a set of theories and methods accepted by the community to guide inquiry. It's a way of thinking. Kuhn describes science as a process involving occasional "revolutions" stemming from crises faced by the dominant theories, followed by periods of "normal science" where the details of the new paradigm are fleshed out. Over time, as the dominant paradigm fails to address an increasing number of important anomalies or challenges, we see a

New York University, Stern School of Business and the Center for Data Science

paradigm shift to a new set of theories and methods – a new way of thinking that better addresses them.

A key feature of paradigms is that they have "exemplars" that guide problem formulation and solution. In physics, for example, the models describing the laws of motion, like Kepler's or Newton's laws of motion, could serve as exemplars that drive hypothesis formulation, observation, and hypothesis testing. In AI, exemplars define the core principles and methods for knowledge extraction, representation, and use. Early approaches favored methods for declaring human-specified knowledge as rules using symbols to describe the world, and an "inference engine" to manipulate the symbols – which was viewed as "reasoning." In contrast, current methods have shifted towards learning more complex statistical representations of the world that are derived almost entirely from data. The latter tend to be better at dealing with the contextual subtleties and complexity that we witness in problems involving language, perception and cognition.

The paradigm shifts in AI have been driven by methods that broke through major walls that were considered to be significant at the time. The first generation of AI research in the late 50s and 60s was dominated by game playing search algorithms (Samuel, 1959, 2000) that led to novel ways for searching various kinds of graph structures. But this type of mechanical search provided limited insight into intelligence, where real-world knowledge seemed to play a major role in solving problems, such as in medical diagnosis and planning. Expert Systems provided a way forward, by representing domain expertise and intuition in the form of explicit rules and relationships that could be invoked by an inference mechanism. But these systems were hard to create and maintain. A knowledge engineer needed to define each relationship manually and consider how it would be invoked in making inferences.

The practical challenges of the knowledge acquisition bottleneck led to the next paradigm shift. As more data became available, researchers developed learning algorithms that could automatically create rules or models directly from the data using mathematical, statistical, or logical, methods, guided by a user-specified objective function.

That's where we are today. Systems such as ChatGPT employ variants of neural networks called Transformers that provide the architecture of large language models (LLMs), which are trained directly from the collection of human expression available on the Internet. They use complex mathematical models with billions of parameters that are estimated from large amounts of publicly available data. While language has been a key area of advancement in recent years, these approaches have been used to enable machines to learn from other modalities of data including vision, sound, smell, and touch. What is particularly important today is the shift from building specialized applications of AI to one where knowledge and intelligence don't have specific boundaries, but transfer seamlessly across applications and to novel situations.

## 2. The Paradigm Shifts in AI

To understand the state of the art of AI and where it is heading, it is important to understand its scientific history, including the bottlenecks that stalled progress in each paradigm and the degree to which they were addressed by each paradigm shift.

Figure 1 sketches out the history of Artificial Intelligence from the Expert Systems era – which spanned the late sixties to the late 80s – to the present.

**Expert Systems and Symbolic AI**

Expert systems are attractive in narrow, well-circumscribed domains in which human expertise is identifiable and definable. They perform well at specific tasks where this expertise can be extracted through interactions with humans, and it is typically represented in terms of relationships among situations and outcomes. The driving force in that paradigm was to apply AI to diagnosis, planning, and design across a number of domains including healthcare, science, engineering, and business. The thinking was that if such systems performed at the level of human experts, they were intelligent.

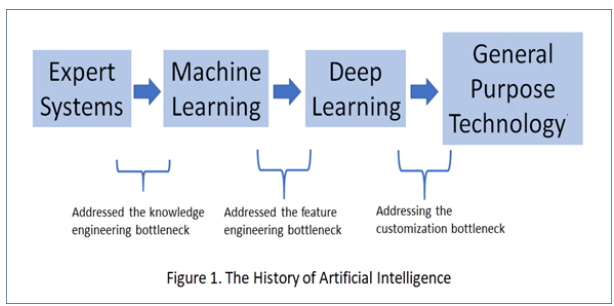

Figure 1. The History of Artificial Intelligence

An early success in medicine was the Internist system (Pople, 1982), which performed diagnosis in the field of internal medicine. Internist represented expert knowledge using causal graphs and hierarchies relating diseases to symptoms. The rule-based expert system Mycin (Buchanan and Shortliffe, 1975) was another early demonstration of diagnostic reasoning involving blood diseases. Other medical applications included the diagnosis of renal failure (Gorry, et.al 1973) and glaucoma (Kulikowski and Weiss 1982).

In addition to applications in medicine, expert systems were also successful in a number of other domains such as engineering (Tzafestas, 1993), accounting (Brown, 1991), tax planning (Shpilberg et.al 1986)., configuration of computer systems (McDermott, 1982), monitoring industrial plants (Nelson, 1982), mineral prospecting (Hart et.al 1978), and identifying new kinds of chemical molecules (Feigenbaum et.al 1970).

The prototypical exemplar for representing knowledge in this paradigm were symbolic relationships expressed in the form of "IF/THEN" rules (Buchanan et.al 1969), "semantic networks," (Quillian, 1968) or structured object representations (Minsky, 1974). But it was difficult to express uncertainty in terms of these representations, let alone combine such uncertainties during inference, which prompted the development of more principled graphical models for representing uncertainty in knowledge using probability theory (Pearl, 1988).

The exemplar was shaped by the existing models of cognition from Psychology, which viewed humans as having a long-term and a short-term memory, and a mechanism for evoking them in a specific context. The knowledge declared by humans in expert systems, such as the rule "*excess bilirubin* → *high pallor*" constituted their long-term memory. An interpreter, also known as the inference engine or "control regime," evoked the rules depending on the context, and updated its short-term memory accordingly. If a patient exhibited unusually high pallor for example, this symptom was noted in short-term memory, and the appropriate rule was evoked from long-term memory to hypothesize its cause, such as "excess bilirubin." In effect, symbolic AI separated the declaration of knowledge from its application.

Research in natural language processing was along similar lines, with researchers seeking to discover the rules of language. The expectation was that once these were fully specified, a machine would follow these rules in order to understand and generate language (Schank, 1990; 1991). This turned out to be exceedingly difficult to achieve.

The major hurdle of this paradigm of top-down knowledge specification was the "knowledge engineering bottleneck." It was challenging to extract reliable knowledge from experts, and equally difficult to represent and combine uncertainty in terms of rules. Collaborations between experts and knowledge engineers could take years or even decades, and the systems became brittle at scale. Furthermore, researchers found that expert systems would often make errors in common-sense reasoning, which seemed intertwined with specialized knowledge. Evaluating such systems was also difficult, if one ever got to that stage. Human reasoning and language seemed much too complex and heterogenous to be captured by top-down specification of relationships. Progress stalled, as the reality, both in research and practice, fell short of expectations.

**Machine Learning**

The supervised machine learning paradigm emerged in the late 80s and 90s, with the maturing of database technology, the emergence of the Internet, and the increasing abundance of observational and transactional data (Breiman, et.al, 1984, Quinlan, 1986). AI thinking shifted away from spoon-feeding highly specified human abstractions to the machine, and towards automatically learning rules from data, guided by human intuition. While symbolic expert systems required humans to specify a model, machine learning enabled the machine to learn the model automatically from curated examples. Model discovery was guided by a "loss function," designed to directly or indirectly to minimize the system's overall prediction error, which by virtue of the data, could be measured in terms of the differences between predictions and empirical reality.

Empirics provided the ground truth for supervision. For example, to learn how to predict pneumonia, also called the *target*, one could collect historical medical records of people with and without pneumonia, intuit and engineer the features that might be associated with the target, and let the machine figure out the relationships from the data to minimize the overall prediction error. Instead of trying to specify the rules, the new generation of algorithms could learn them from data using optimization. Many such algorithms emerged, but the common thread among them was that they belonged to the broad class of "function approximation" methods that used data and a user-defined objective function to guide knowledge discovery.

This shift in perspective transformed the machine into a generator and tester of hypotheses that used optimization – the loss function – to focus knowledge discovery. This ability made machines capable of automated inquiry without a human in the loop. Instead of a being a passive repository of knowledge, the machine became an active "what if" explorer, capable of asking and evaluating its own questions. This enabled data-driven scientific discovery (Hey et.al, 2009).

The epistemic criterion in machine learning for something to count as knowledge was accurate *prediction* (Popper, 1963; Dhar, 2013). This conforms to Popper's view of using the predictive power theories as a measure of their goodness. Popper argued that theories that sought only to explain a phenomenon were weaker than those that made "bold" *ex-ante* predictions" that were objectively falsifiable. Good theories stood the test of time. In his 1963 treatise on this subject, *Conjectures and Refutations*, Popper characterized Einstein's theory of relativity as a "good" one, since it made bold predictions that can be falsified easily, yet all attempts at falsification of the theory have failed.

The exemplars for supervised machine learning are relationships derived from data that is specified in (X,y) pairs, where "y" are data about the target to be predicted based on a situation described by the vector of observable features "X." This exemplar has a very general form: the discovered relationships can be "IF/THEN" rules, graph structures such as Bayesian networks (Pearl, 1988; Blei et.al, 2003), or implicit mathematical functions expressed via weights in a neural network (Rosenblatt, 1958; Hopfield, 1982). Once learned, this knowledge could be viewed as analogous to memory, invoked depending on context, and updatable over time.

But there's no free lunch with machine learning. There is a loss of transparency in what the machine has learned. Neural networks, including large language models, are particularly opaque in that it is difficult to assign meanings to the connections among neurons, let alone combinations of them. Even more significantly, the machine learning paradigm introduced a new bottleneck, namely, requiring the curation of available data using some sort of vocabulary that the machine can understand. This required that the *right features* be created from the raw data. For example, to include an MRI image as input into the diagnostic reasoning process, the contents of the image had to be expressed in terms of features of the vocabulary such as "inflammation" and "large spots on the liver." Similarly, a physician's notes about a case had to be condensed into features that the machine could process. This type of *feature engineering* was cumbersome. Specifying the labels accurately could also be costly and time-consuming. These were major bottlenecks for the paradigm.

What was direly needed was the ability of the machine to deal directly with the raw data emanating from the real world, instead of relying on humans to perform the often difficult translation of feature engineering. Machines needed to ingest raw data such as numbers, images, notes or sounds directly, ideally without curation by humans.

## Deep Learning

The next AI paradigm, "Deep learning," made a big dent in the feature engineering bottleneck by providing a solution to perception, such as seeing, reading, and hearing. Instead of requiring humans to describe the world for the machine, this generation of algorithms could consume the raw input similar to what humans use, in the form of images, language, and sound. "Deep neural nets" (DNNs), which involve multiple stacked layers of neurons, form the foundation of vision and language models (Hinton, 1992; LeCun and Bengio, 1998). While learning still involves adjusting the weights among the neurons, the "deep" part of the neural architecture is important in translating the raw sensory input automatically into machine-computable data.

The exemplar in deep learning is a multi-level neural network architecture. Adjusting the weights among the neurons makes it a *universal function approximator* (Cybenko, 1989), where the machine can approximate any function, regardless of its complexity, to an acceptable degree of precision. What is unique about DNNs is the organization of hidden layers between the input and output that *learn the features* implicit in the raw data instead of requiring that they be specified by humans. A vision system, for example, might learn to recognize features common to all images, such as lines, curves and colors from the raw images that make up its training data. These can be combined variously to make up more complex image parts such as windows, doors and street signs that are represented by "downstream" layers of the deep neural network. In other words, the DNN tends to have an organization, where more abstract and latent concepts that are closer to its output are composed from more basic features represented in the layers that are closer to the input.

The same ideas have been applied to large language models (LLMs) from which systems like ChatGPT are built. They learn the implicit relationships among things in the world from large amounts of text from books, magazines, web-posts etc. As in vision, we would expect layers of the neural network that are closer to the output to represent more abstract concepts, relative to layers that are closer to the input. However, we don't currently understand how DNNs organize and use such knowledge, or how they represent relationships in general. This depends on what they are trained for.

In language modeling, for example, the core learning task is typically to predict the next occurrence of an input sequence. This requires a considerable amount of knowledge and understanding of the relationships among the different parts of the input. Large language models use a special configuration of the "Transformer" neural architecture, which represents language as a contextualized sequence, where context is represented by estimating dependencies between each pair of the input sequence (Vaswani et.al 2017). Because this pairwise computation grows sharply with the length of the input, engineering considerations constrain the length of the input sequences – its span of attention, for which LLMs are able to maintain context.

This Transformer architecture holds both long term memory, represented by the connections between neurons, as well as the context of a conversation – the equivalent of short-term memory – using its "attention mechanism," which captures the relationships between all parts of the input. For example, it is able to tell what the pronoun "it" refers to in the sentences "The chicken didn't cross the road because it was wet" and "The chicken didn't cross the road because it was tired." While humans find such reasoning easy by invoking common-sense, previous paradigms failed at such kinds of tasks that require understanding context. The architecture also works remarkably well in vision, where it is able to capture the correlation structure between the various parts of an image.

The downside is that DNNs are large and complex. What pre-trained language models learn as a by-product of learning sequence prediction is unclear because their knowledge – the meanings and relationships among things – is represented in a "distributed" way, in the form of weighted connections among the layers of neurons. In contrast to Expert Systems, where relationships are specified in "localized" self-contained chunks, the relationships in a DNN are smeared across the weights in the network and much harder to interpret.

Nevertheless, the complexity of the neural network architecture – typically measured by the number of layers and connections in the neural network (its parameters) – is what allows the machine to recognize context and nuance. It is remarkable that the pre-trained LLM can be used to explain why a joke is funny, summarize or interpret a legal document, answer questions, and all kinds of other things that is wasn't explicitly trained to do.

Bowman (2023) conjectures that the autocomplete task was serendipitous: it was just at the right level of difficulty, where doing well conversationally forced the machine to learn a large number of other things about the world. In other words, a sufficiently deep understanding about the world, including common-sense, is *necessary* for language fluency. However, current-day machines can't match humans in terms of common sense. As of this writing, ChatGPT fails at the Winograd Schema task (Winograd, 1972), which involves resolving an ambiguous pronoun in a sentence. For example, when asked what the "it" refers to in the sentence "the trophy wouldn't fit into the suitcase because it was too small," ChatGPT thinks that the "it" refers to the trophy. The right answer requires the use of common-sense, and cannot be determined by structure alone.

These are the challenges for the new and still emerging paradigm of AI, namely one of *General Intelligence*, where expertise and common-sense can blend together more seamlessly, as can different modalities of information.

Table 1 summarizes the properties of each paradigm in terms of how knowledge is acquired (its source), the exemplar that guides problem formulation, its capability, and the degree to which the input is curated. The "+" prefix means "in addition to the previous case."

|  | Knowledge Source | Exemplar | Capability | Data Curation |
|---|---|---|---|---|
| **Expert Systems** | **Human** | **Rules** | **Follows** | **High** |
| **Machine Learning** | **+ Databases** | **Rules/Networks** | **+ Discovers Relationships** | **Medium** |
| **Deep Learning** | **+ Sensory** | **Deep Neural Networks** | **+ Senses Relationships** | **Low** |
| **General Intelligence** | **+ Everything** | **Pre-Trained Deep Neural Networks** | **+ Understands the World** | **Minimal** |

**Table 1: The Paradigm Shifts in AI**

## 3. General Intelligence

Pre-trained models are the foundation for the General Intelligence paradigm. Previous AI applications were tuned to a task. In order to predict pneumonia in a hospital, for example, the AI model was trained using cases from that hospital alone, and wouldn't necessarily transfer to a nearby hospital, let alone a different country. In contrast, General Intelligence is about the ability to integrate knowledge about pneumonia with other diseases, conditions, geographies, etc., like humans are able to do, and to apply the knowledge to unforeseen situations. In other words, General Intelligence refers to an integrated set of essential mental skills that include spatial, numerical, mechanical, verbal, reasoning, and common sense abilities, which underpin performance on *all* mental tasks (Cattell, 1963). Such knowledge is easily *transferrable* across tasks, and can be applied to novel situations.

Each paradigm shift greatly expanded the scope of applications. Machine Learning brought structured databases to life. Deep Learning went further, enabling the machine to deal with structured and unstructured data about an application directly from the real world, as humans are able to do.

Pre-trained models provide the building blocks for General Intelligence by virtue of being domain-independent, requiring minimal curation,[2] and being transferrable across applications.

The shift to pre-trained models represents a fundamental departure from the previous paradigms, where knowledge was carefully extracted and represented. AI was an application, and tacit knowledge and common-sense reasoning were add-ons that were separate from expertise. The CYC project (Lenat et.al 1990) was the first major effort to explicitly teach the machine common sense. It didn't work as the designers had hoped. There's too much tacit knowledge and common sense in human interaction that is evoked depending on context, and intelligence is much too complex and heterogenous to be compartmentalized and specified in the form of rules.

In contrast, pre-trained models eschew boundaries, as in the pneumonia example. Rather, they integrate specialized and general knowledge, including data about peoples' experiences across a range of subjects. Much of this type of knowledge became available because of the Internet, where in the short span of a few decades, humanity expressed thousands of years of its history in terms of language, along with social media and conversational data on a wide array of subjects. All humans became potential publishers and curators, providing the training data for AI to learn how to communicate fluently. Hinton describes large language models like ChatGPT akin to an alien species that has enthralled us because they speak such good English.

---

[2] The data curation in pre-trained LLMs like GPT-3 is primarily in the choice of sources and tokenization. AI Systems like ChatGPT3 use additional conversational training and RLHF (reinforcement learning with Human Feedback) to curate their response to be socially acceptable.

It is important to appreciate that in learning to communicate in natural language, AI has broken through two fundamental bottlenecks simultaneously. First, we are now able to communicate with machines on *our* terms. This required solving a related problem, of integrating and transferring knowledge about the world, including common sense, seamlessly into a conversation about any subject. Achieving this capability has required the machine to acquire the various types of knowledge simultaneously – expertise, common sense, and tacit knowledge – all of which are embedded in language. Things are connected in subtle ways, which provides the basis for "meaning" and "understanding," which AI pioneer Marvin Minsky describes in terms of "associations" and "perspectives:"

*What is the difference between merely knowing (or remembering, or memorizing) and understanding? We all agree that to understand something, we must know what it means, and that is about as far as we ever get. A thing or idea seems meaningful only when we have several different ways to represent it–different perspectives and different associations. Then we can turn it around in our minds, so to speak: however it seems at the moment, we can see it another way and we never come to a full stop. In other words, we can 'think' about it. If there were only one way to represent this thing or idea, we would not call this representation thinking.*(Minsky, 1981)

Conversational agents such as ChatGPT display a remarkable ability to adapt and combine contexts in maintaining conversational coherence, This capability, where the machine can understand what we are saying well enough to maintain a conversation, enables a new kind of interaction, where the machine is able to acquire high quality training data seamlessly "from the wild" and learn in parallel with its operation.

As in deep learning, the exemplar in General Intelligence paradigm is the deep neural network, whose properties we now trying to understand, along with the general principles that underpin their performance. One such principle in the area of LLMs is that performance improves by increasing model complexity, data size, and compute power across a wide range of tasks (Kaplan, et.al, 2020). These "scaling laws of AI" indicate that predictive accuracy on the autocompletion task improves with increased compute power, model complexity, and data. If this measure of performance on autocompletion is a good proxy for General Intelligence, the scaling laws predict that LLMs should continue to improve with increases in compute power and data. A related phenomenon to performance improvement with scaling may be the "emergence" of *new abilities* at certain tipping points of model size (Wei, et.al, 2022) that don't exist at smaller model sizes.

At the moment, there are no obvious limits to these dimensions in the development of General Intelligence. On the data front, for example, in addition to additional language data that will be generated by humans on the Internet, other modalities of data such as video are now becoming more widely available. Indeed, a fertile area of research is how machines will *integrate* data from across multiple sensory modalities including vision, touch and smell, like humans are able to do. In short, we are in the early innings of the new paradigm, where we should see continued improvement of pre-trained models and General Intelligence with increases in the volume and variety of data and computing power. However, this should not distract us from the fact that several fundamental aspects of intelligence are still mysterious, and unlikely to be answered solely by making existing models larger and more complex.

Nevertheless, it is worth noting that the DNN exemplar of General Intelligence has been adopted by a number of disciplines including psychology, neuroscience, linguistics, and philosophy, that seek to *explain* intelligence, meaning, and understanding. This has arguably made AI more interdisciplinary by unifying its theses with its engineering and design perspectives. Explaining and understanding the behavior of DNNs in terms of a set of core principles of its underlying disciplines is an active area of research in the current paradigm.

The progression towards the General Intelligence has followed a path of increasing scope of machine intelligence. The first paradigm was "Learn from humans." The next one was "Learn from curated data." This was followed by "Learn from any kind of data." The current paradigm is "Learn from all kinds of data it in a way that transfers to novel situations." The latest paradigm shift makes AI a general purpose technology and a commodity, one that should keep improving in terms of quality with increasing amounts of data and computing power.

## 4. AI as a General-Purpose Technology

Paradigm shifts as defined by Kuhn are followed by periods of "normal science," where the details of the new paradigm are fleshed out. We are in the early stages of one such period.

Despite their current limitations, pre-trained LLMs and conversational AI have unleashed applications in language and vision, ranging from support services that require conversational expertise to creative tasks such as creating documents or videos. As the capability provided by these pre-trained models grows and becomes embedded in a broad range of industries and functions, AI is transitioning from a bespoke set of tools to a "General Purpose Technology," from which applications are assembled. Like electricity, intelligence becomes a commodity.

Economists use the term *general-purpose technology* –of which electricity and the Internet are examples – as a new method for producing and inventing that is important enough to have a protracted aggregate economic impact across the economy (Jovanovic and Rousseau, 2005).

Bresnahan and Trachtenburg (1995) describe general purpose technologies in terms of three defining properties:

"**pervasiveness** *– they are used as inputs by many downstream sectors),*

inherent **potential for technical improvements**, *and*

**innovational complementarities** *– the productivity of R&D in downstream sectors multiplies as a consequence of innovation in the general purpose technology, creating productivity gains throughout the economy."*

How well does the General Intelligence paradigm of AI meet these criteria?

Arguably, AI is already pervasive, embedded increasingly in applications without our realization. And with the new high bandwidth human-machine interfaces enabled by conversational AI, the quality and volume of training data that machines like ChatGPT can now acquire as they operate is unprecedented. Other sensory data from video and other sources will continue to lead to improvements in pre-trained models and their downstream applications.

The last of the three properties, innovation complementarities, may take time to play out at the level of the economy. With previous technologies such as electricity and IT, growth rates were *below* those attained in the decades immediately preceding their arrival (Jovanovic and Rousseau, 2005). This phenomenon was also observed by the economist Robert Solow who famously commented that "IT was everywhere except in the productivity statistics." (Solow, 1987). Erik Brynolffson and his colleagues subsequently explained Solow's observation in terms of the substantial complementary investments required to realize the benefits of IT (Brynjolfsson et.al, 2018), where productivity emerged after a significant lag. With electricity, for example, it took decades for society to realize its benefits, since motors needed to be replaced, factories needed redesign, and workforces needed to be reskilled. IT was similar, as was the Internet.

AI is similarly in its early stages, where businesses are scrambling to reorganize business processes and rethinking the future of work. Just as electricity required the creation of an electric grid and the redesign of factories, AI will similarly require a redesign of business processes in order to realize productivity gains from this new general purpose technology (Brynjolfsson et.al 2023). Such improvements take time to play out, and depend on effective complementary investments in processes and technologies.

## 5. Challenges of Current Paradigm: Trust and Law

We should not assume that we have converged on the "right paradigm" for AI. The current paradigm will undoubtedly give way to one that addresses its shortcomings.

Indeed, paradigm shifts do not always improve on previous paradigms in every way, especially in their early stages, and the current paradigm is no exception. New theories often face resistance and challenges initially, while their details are being filled in (Laudan, 1978). For example, the Copernican revolution faced numerous challenges in explaining certain recorded planetary movements that were explained by the existing theory, until new methods and measurements emerged that provided strong support for the new theory (Kuhn, 1956).

Despite the current optimism about AI, the current paradigm faces a serious challenge of trust, that stems in large part from its representation of knowledge that is opaque to humans. Systems such as ChatGPT can be trained on orders of magnitude more cases than a human expert will encounter in their lifetime, but their ability to explain themselves and introspect is very limited relative to humans. And we can never be sure that they are correct, and not "hallucinating," that is, filling in their knowledge gaps with answers that look credible but are incorrect. It's like talking to someone intelligent that you can't always trust.

These problems will need to be addressed if we are to trust AI. Since the data for pre-trained models are not curated, they pick up on the falsehoods, biases, and noise in their training data. Systems using LLMs can also be unpredictable, and systems based on them can exhibit racist or other kinds of undesirable social behavior that their designers didn't intend. While designers might take great care to prohibit undesirable behavior via training using "reinforcement learning via human feedback" (RLHF), such guardrails don't always work as intended. The machine is relatively inscrutable.

Making AI explainable and truthful is a big challenge. At the moment, it isn't obvious whether this problem is addressable solely by the existing paradigm, whether it will require a new paradigm, or whether it will be addressed via an integration of the symbolic and neural approaches to computation.

The unpredictability of AI systems built on pre-trained models also poses new problems for trust. The output of LLM-based AI systems on the same input can vary, a behavior we associate with humans but not machines (Dhar, 2022). To the contrary, we expect machines to be deterministic, not "noisy" or inconsistent like humans. Until now, we have expected consistency from machines.

While we might consider the machine's variance in decision-making as an indication of creativity

– a human-like behavior – it poses severe risks, especially when combined with its inscrutability and an uncanny ability to mimic humans. Machines are already able to create "deep fakes" which can be undistinguishable from human creations. We are seeing the emergence of things like fake pornography, art, and documents. It is exceedingly difficult to detect plagiarism, or to even define plagiarism or intellectual property theft, given the large corpus of public information on which LLMs have been trained. When will such risks lie with the creators of pre-trained models, applications that use them, or their users? Existing laws are not designed to address such problems, and will need to be expanded to recognize them, to limits their risks, and specify culpability.

Finally, inscrutability also creates a larger, existential risk to humanity, which could become a crisis for the current paradigm. For example, in trying to achieve goals that we give the AI, such as "save the planet," we have no idea about the sub-goals the machine will create in order to achieve its larger goals. This is known as "the alignment problem," in that it is impossible to determine whether the machine's hidden goals are aligned with ours. In saving the planet, for example, the AI might determine that humans pose the greatest risk to its survival, and hence they should be contained or eliminated (Bostrom 2014; Russell, 2019, Christian, 2020).

So, even as we celebrate AI as a technology that will have far-reaching impacts on society, economics, and humanity – potentially exceeding that of other general purpose technologies such as electric power and the Internet – trust and alignment remain disconcertingly unaddressed. They are the most pressing ones that humanity faces today.

**References**

1. Bostrom, N., SuperIntelligence, Oxford University Press, 2014.
2. Bowman, S.R., et.al. Measuring Progress on Scalable Oversight for Large Language Models, November 2022. https://arxiv.org/abs/2211.03540
3. Brynjolfsson, E., Daniel Rock, D., and Syverson, D., Unpacking the AI-Productivity Paradox, Sloan Management Review, January 2018. https://sloanreview.mit.edu/article/unpacking-the-ai-productivity-paradox/?utm_source=twitter&utm_medium=social&utm_campaign=sm-direct
4. Brynjolfsson, E., Li, Danielle, Raymond, L., Generative AI at Work, https://arxiv.org/abs/2304.11771 2023
5. Brown, C.E., Expert systems in public accounting: Current practice and future directions, Expert Systems with Applications, Volume 3, Issue 1, 1991, Pages 3-18, ISSN 0957-4174, https://doi.org/10.1016/0957-4174(91)90084-R.
6. Caruana, R., Lou, Y., Gehrke, J., Koch, P., Sturm, M., Elhadad, N., Intelligible Models for HealthCare: Predicting Pneumonia Risk and Hospital 30-day Readmission, KDD 2015, Sydney, Australia.
7. Chalmers, D., Reality+: Virtual Worlds and the Problems of Philosophy. W.W. Norton & Co, 2022.
8. Charniak, E., and McDermott, D., Introduction to Artificial Intelligence, Addison-Wesley, 1985.
9. Christian, B., The Alignment Problem: Machine Learning and Human Values, Brilliance Publishing, 2020.
10. Cybenko, G. (1989). "Approximation by superpositions of a sigmoidal function". *Mathematics of Control, Signals, and Systems*. **2** (4): 303–314.
11. Dhar, V., Data science and prediction, Communications of the ACM, Volume 56, Issue 12, December 2013.
12. Dhar, V., Bias and Noise in Humans and AI, Journal of Investment Management, vol 20, number 4, December 2022.

13. Dhar, V., When to Trust Machines With Decisions and When Not To, Harvard Business Review, May 2016.
14. Doersch, C; Zisserman, A. *"Multi-task Self-Supervised Visual Learning"*. 2017 IEEE International Conference on Computer Vision (ICCV). IEEE: 2070–2079. arXiv:1708.07860. doi:10.1109/iccv.2017.226. ISBN 978-1-5386-1032-9. S2CID 473729.
15. Feigenbaum, Edward & Buchanan, Bruce & Lederberg, Joshua. (1970). On generality and problem solving: A case study using the DENDRAL program. Machine Intelligence. 6.
16. Hart, P.E., Duda, R.O. & Einaudi, M.T. PROSPECTOR—A computer-based consultation system for mineral exploration. *Mathematical Geology* **10**, 589–610 (1978). https://doi.org/10.1007/BF02461988
17. Hey, T., Tansley, S., Tolle, K., Gray, J., The Fourth Paradigm: Data-Intensive Scientific Discovery, 2009. https://www.microsoft.com/en-us/research/wp-content/uploads/2009/10/Fourth_Paradigm.pdf
18. Gorry GA, Kassirer JP, Essig A, and Schwartz WB (1973). "Decision analysis as the basis for computer-aided management of acute renal failure". The American Journal of Medicine. **55** (4): 473–484. doi:10.1016/0002-9343(73)90204-0. PMID 4582702.
19. Hart,P., Duda, R., Einaudi, M., PROSPECTOR—A computer-based consultation system for mineral exploration, Journal of the International Association for Mathematical Geology volume 10, number 5, 1978. https://doi.org/10.1007/BF02461988
20. Hinton, G., How neural networks learn from experience. Scientific American, September 1992.
21. Hopfield, J. J. (1982). Neural networks and physical systems with emergent collective computational abilities, Proceedings of the National Academy of Sciences. **79** (8): 2554–2558.
22. Jovanovic, B., and Rousseau, P., General Purpose Technologies, in *Handbook of Economic Growth, Volume 1B*. Edited by Philippe Aghion and Steven N. Durlauf, 2005.
23. Kaplan, L., et.al 2020. Scaling Laws for Neural Language Models https://arxiv.org/pdf/2001.08361.pdf
24. Kumar, A., Irsoy, O., Ondruska, P., Iyyer, M., Bradbury, J., Gulrajani, I.,Zhong, V., Paulus, R., Socher, R., Ask me anything: Dynamic memory networks for natural language processing. In *International conference on machine learning* (pp. 1378-1387).
25. Kuhn, Thomas S. The Structure of Scientific Revolutions. University of Chicago Press, 1962.
26. Kuhn, Thomas S. The Copernican Revolution. Harvard Press, 1956.
27. Kulikowski, C. A. and Weiss, S. M.  "Representation of Expert Knowledge for Consultation: The CASNET and EXPERT Projects."  Chapter 2 in Szolovits, P. (Ed.) *Artificial Intelligence in Medicine*. Westview Press, Boulder, Colorado.  1982.
28. Laudan, L., Progress and Its Problems, University of California Press 1978.
29. LeCun, Y., and Bengio, Y., Convolutional networks for images, speech, and time series, in The handbook of brain theory and neural networks, October 1998 Pages 255–258 https://dl.acm.org/doi/10.5555/303568.303704
30. *L*enat, Douglas B.; Guha, R. V.; Pittman, Karen; Pratt, Dexter; Shepherd, Mary (August 1990). "Cyc: Toward Programs with Common Sense". Commun. ACM. **33** (8): 30–49.
31. McDermott, J., R1: a rule-based configurer of computer systems, Artificial Intelligence, Volume 19, Issue

1, September 1982 pp 39–88 https://doi.org/10.1016/0004-3702(82)90021-2
32. Meng,, K., Sharma, A., Andonian, A. Belinkov, Y., Bau, D., 2022. https://arxiv.org/pdf/2210.07229.pdf
33. Minsky, M., A Framework for Representing Knowledge, in *The Psychology of Computer Vision*, P. Winston (Ed.), McGraw-Hill, 1975.
34. Minsky, M., Music, Mind, and Meaning, Computer Music Journal, Fall 1981
35. Pople, H. E (1982), Heuristic Methods for Imposing Structure on Ill-Structured Problems, in Szolovits, P. (Ed.) *Artificial Intelligence in Medicine*. Westview Press, Boulder, Colorado. 1982.
36. Nelson, W. R. (*1982). "REACTOR: An Expert System for Diagnosis and Treatment of Nuclear Reactors".*
37. *Ouyang, L,. et.al, 2022. Training Models to Follow Instructions With Human Feedback.,* https://arxiv.org/pdf/2203.02155.pdf
38. Pearl, J., Probabilistic Reasoning in Intelligent Systems, Morgan Kauffman Publishers, 1988.
39. Quillian, J.R., Semantic Networks, In Marvin L. Minsky (ed.), *Semantic Information Processing*. MIT Press (1968)
40. Quinlan, J. R. 1986. Induction of Decision Trees. *Machine Learning* volume 1, No1 (Mar. 1986), 81–106
41. Rosenblatt, F. (1958). "The Perceptron: A Probabilistic Model For Information Storage And Organization In The Brain". Psychological Review. **65** (6): 386–408. CiteSeerX 10.1.1.588.3775
42. Russell, Stuart., Human Compatible: AI and the Problem of Control, Penguin, 2019.
43. Samuel, Arthur L. (1959). *"Some Studies in Machine Learning Using the Game of Checkers". IBM Journal of Research and Development.* **44**: 206–226.
44. Samuel, A. L. (2000). *"Some studies in machine learning using the game of checkers". IBM Journal of Research and Development. IBM.* **44**: 206–226. doi:10.1147/rd.441.0206
45. Schaeffer, R. Miranda, B., and Koyejo, S., (2023) Are Emergent Properties of Large Language Models a Mirage? https://arxiv.org/abs/2304.15004
46. Schank, Roger. *The Connoisseur's Guide to the Mind: How we think, How we learn, and what it means to be intelligent*. Summit Books, 1991.
47. Schank, Roger. *Tell Me A Story: A new look at real and artificial memory*. Scribner's, 1990
48. Shanahan, M., Talking About Large Language Models. https://arxiv.org/pdf/2212.03551.pdf
49. Shortliffe EH, and Buchanan BG (1975). *"A model of inexact reasoning in medicine". Mathematical Biosciences.* **23** *(3–4): 351–379.*
50. Shpilberg,D., Graham, L., Schatz, H., ExperTAX$^{sm}$: an expert system for corporate tax planning, Wiley, July 1986 https://doi.org/10.1111/j.1468-0394.1986.tb00487.x
51. *Szolovits P, Patil RS, and Schwartz WB . (1988). "Artificial intelligence in medical diagnosis". Annals of Internal Medicine.* **108** *(1): 80–87.*
52. *Tzafestas, S., Expert Systems in Engineering Applications, Springer-Verlag, 1993.*
53. Vaswani, A., Shazeer, N., Parmar, N., Uszkoreit, J., Jones, L., Gomez, A., Kaiser, L. Polosukhin, I., (2017) Attention is All You Need., https://arxiv.org/abs/1706.03762
54. Wei, J., Tay, Y., Bommasini, R., Raffel, C., Zoph, B., Borgeaud, S., Y, D., ., Emergent Properties of Large Language Models, 2022. https://arxiv.org/pdf/2206.07682.pdf

55. Wei, J., Wang, X., Schuurmans, D., Bosma M., Ichter, B., Xia, F., Chi, E., Li, Q., Zhou, D., 2023. https://arxiv.org/pdf/2201.11903.pdf
56. Winograd, Terry (January 1972). Understanding Natural Language. Cognitive Psychology. **3** (1): 1–191. doi:10.1016/0010-0285(72)90002-3